%% file: IEEE-Main.tex
\documentclass[conference]{IEEEtran}
\IEEEoverridecommandlockouts
\usepackage{cite}
\usepackage{amsmath,amssymb,amsfonts}
\usepackage{algorithmic}
\usepackage{graphicx}
\usepackage{textcomp}
\usepackage{xcolor}
\usepackage[utf8]{inputenc}
\usepackage{verbatim}
\usepackage{url}
\usepackage{booktabs}
\usepackage{float}
\def\BibTeX{{\rm B\kern-.05em{\sc i\kern-.025em b}\kern-.08em
    T\kern-.1667em\lower.7ex\hbox{E}\kern-.125emX}}

\usepackage{newunicodechar}
\newunicodechar{−}{\textminus}

\begin{document}

\title{Do Methods Support the Claims? Intra-Paper Verification for Peer Review\\
}

%\author{
%\IEEEauthorblockN{Ranjitha Shivaprasad Ballakuraya, Arash Mahyari, Ashok Srinivasan}
%, Brent Venable}
%\IEEEauthorblockA{University of West Florida, Florida, USA\\
%Email: rb194@students.uwf.edu, amahyari@uwf.edu, asrinivasan@uwf.edu} %bvenable@uwf.edu}
%}

\author{
\IEEEauthorblockN{Ranjitha Shivaprasad Ballakuraya}
\IEEEauthorblockA{
\textit{Dept. of Intelligent Systems and Robotics} \\
\textit{University of West Florida} \\
Pensacola, FL, USA \\
rb194@students.uwf.edu
}
\and
\IEEEauthorblockN{Arash Mahyari}
\IEEEauthorblockA{
\textit{Dept. of Intelligent Systems and Robotics} \\
\textit{University of West Florida} \\
Pensacola, FL, USA \\
amahyari@uwf.edu
}
\and
\IEEEauthorblockN{Ashok Srinivasan}
\IEEEauthorblockA{
\textit{Dept. of Computer Science} \\
\textit{University of West Florida} \\
Pensacola, FL, USA \\
asrinivasan@uwf.edu
}
}
\maketitle
\begin{abstract}
The growing volume of scientific submissions has motivated interest in using large language models (LLMs) to assist peer review. Existing automated novelty-assessment approaches typically compare a paper’s claimed contributions against prior literature, implicitly assuming that these contributions are accurately realized in the work itself. Human reviewers, however, frequently challenge novelty claims not because similar ideas already exist, but because the methodological evidence presented in the paper does not adequately support them. This internal mismatch between claimed contributions and methodological realization is rarely examined by current LLM-based review systems. To address this gap, we introduce intra-paper claim verification, a framework that evaluates whether novelty claims articulated in a paper are substantiated by the methods used to realize them. The framework employs an LLM to extract novelty claims from the introduction, retrieve claim-relevant methodological evidence, and assess whether the methods substantiate the stated contributions. Assessment is guided by reviewer-inspired evaluation criteria derived inductively from human peer reviews collected from 182 ICLR 2025 papers. These criteria capture recurring reviewer concerns related to novelty, methodology, clarity, and other issues and are used to generate structured reviewer-style assessments of claim substantiation.

We evaluate the framework by comparing LLM-generated review comments against human reviewer concerns on a balanced subset of accepted and rejected papers. Human evaluation demonstrates significant alignment between framework-generated assessments and human reviewer concerns, particularly for novelty-related issues ($p < 0.001$, Cohen's $d = 1.17$). BERTScore further distinguishes corresponding human--LLM review pairs from mismatched controls, indicating that the framework captures concerns consistent with human reviewer observations. These results indicate that intra-paper claim verification offers a valuable complement to literature-based novelty assessment by evaluating whether claimed contributions are substantiated by methodological evidence, thereby supporting more reliable LLM-assisted peer review.

The source code, prompts, evaluation materials, and supporting data used in this study are publicly available.\footnote{\url{https://github.com/Ranjitha2493/intra-paper-claim-method-verification}}

\begin{comment}
    
Scientific peer review faces scalability challenges due to rising submission volumes and limited expert availability, leading to reviewer overload and inconsistent evaluations. Large language models (LLMs) offer promising support but have yet to fully address the verification of novelty claims within manuscripts. Here we introduce intra-paper claim verification, a novel task that systematically assesses whetter a paper's novelty claims in the introduction are substantiated by its methodological contributions. Utilizing a structures GPT 4o-based pipeline, we extract claims, retrieve relevant methods, and generate structured reviews for 182 ICLR 2025 papers, comparing theses to human reviews across four dimensions: novelty, methodology, clarity and other concerns. Our findings reveal meaningful alignment between LLM-generated and human assessments, particularly for novelty-related concerns. This approach provides a scalable, evidence-based tool to assist reviewers in identifying inconsistency between claimed contributions and methodological support, potentially enhancing the efficiency and consistency of scientific peer review while preserving human oversight.
\end{comment}
\end{abstract}


%%%%%%%%%%%%%%%%%%%%%%%%%%%%%%%%%%%%%%%%%%%%%%%%%%%%%%%%%%%%%%%%%%%%%%%%%%%%%%%%%%%%%%%%%%%
%%%%%%%%%%%%%%%%%%%%%%%%%%%%%%%%%%%%%%%%%%%%%%%%%%%%%%%%%%%%%%%%%%%%%%%%%%%%%%%%%%%%%%%%%%%
\input IEEE-Introduction.tex

\input IEEE-RelatedWork.tex

\input IEEE-Methodology.tex

\input IEEE-Evaluation.tex
\input IEEE-Result.tex

\input IEEE-LimitationConclusion.tex

%%%%%%%%%%%%%%%%%%%%%%%%%%%%%%%%%%%%%%%%%%%%%%%%%%%%%%%%%%%%%%%%%%%%%%%%%%%%%%%%%%%%%%%%%%%
%%%%%%%%%%%%%%%%%%%%%%%%%%%%%%%%%%%%%%%%%%%%%%%%%%%%%%%%%%%%%%%%%%%%%%%%%%%%%%%%%%%%%%%%%%%

\bibliographystyle{IEEEtran}
\input{IEEE-Main.bbl}

\end{document}

%% file: IEEE-Introduction.tex
\section{Introduction}

The academic community is facing a scalability challenge that undermines the integrity of scientific publishing. Submission volumes at leading venues have grown exponentially over the past decade, outpacing the availability of qualified domain experts~\cite{hanson2024strain}. This structural imbalance has contributed to reviewer overload, reviewer scarcity, prolonged evaluation cycles, and concerns regarding the consistency and sustainability of peer review~\cite{aczel2025present,gupta2025peer}. Prior studies have further demonstrated substantial variability in peer-review outcomes, highlighting the inherent subjectivity of the review process~\cite{lawrence2022neurips}. 

Concurrently, the emergence of LLMs has introduced new possibilities for augmenting the peer review process. LLMs have demonstrated proficiency in comprehending and critiquing scientific discourse, prompting experimental deployments as supplementary reviewers at several prominent venues~\cite{jin2024agentreview, biswas2026ai}. However, the degree to which LLM-generated reviews align with expert human judgment remains an open question, particularly across the critical dimensions that reviewers emphasize.

Reviewers commonly request stronger validation, clearer methodological justification, broader contextualization with prior work, and a more balanced discussion of limitations versus claims~\cite{s43588-026-00989-9}. Publicly available meta-reviews on platforms such as OpenReview~\cite{openreview} further highlight recurring concerns regarding overstated novelty, partially realized contributions, and insufficient empirical evidence. While these observations underscore the necessity of deep scientific reasoning and contextual judgment, AI systems can effectively support large-scale evidence synthesis by processing vast volumes of manuscripts and organizing information consistently. Consequently, AI is best conceptualized as a complementary tool designed to augment expert workflows rather than replace human peer reviewers~\cite{sarkar2026ai}.

Among the evaluative dimensions of peer review, novelty remains a primary determinant of acceptance. Prior work on LLM-based review automation has largely approached novelty assessment through literature-grounded comparison—evaluating the extent to which a paper's contributions differ from existing research using Retrieval-Augmented Generation (RAG)~\cite{afzal-etal-2026-beyond, moussa2025scholareval}. However, expert critiques often extend beyond external comparisons, frequently questioning whether the innovations claimed within the paper are adequately realized and supported by the internal methodological evidence. 

To address this gap, this study introduces \textit{intra-paper claim verification}: a claim-level evaluation task that assesses whether the novelty claims articulated in a manuscript's Introduction are fully substantiated by the empirical evidence described in its Methods section. Unlike existing LLM-based approaches that evaluate manuscripts holistically or against external literature corpora, our proposed framework focuses entirely on the internal consistency of the manuscript.

We present a structured pipeline in which an advanced frontier LLM (specifically, GPT-4o)~\cite{hurst2024gpt} operationalizes intra-paper claim verification through four sequential stages. First, novelty claims are extracted from the Introduction. Second,human reviews-derived evaluation categories are identified from human peer reviews. Third methodological evidence associated with each claim is analyzed to generate structured LLM review assessments, while human reviewer comments are extracted and categorized according to the same evaluation categories. Finally, both assessments are summarized for evaluation.

The proposed framework is validated using 20 randomly selected papers from a corpus of 182 ICLR 2025 submissions, spanning both accepted and rejected decisions. While the full corpus is used to derive the reviewer-informed evaluation categories, the primary validation relies on a blinded expert human study. Four scientist evaluators independently assess the extent to which framework-generated reviews align with concerns raised by human reviewers. To minimize bias, both framework-generated assessments and control reviews generated for papers outside the evaluation dataset are anonymized and presented without revealing their source. This design enables a direct examination of whether the framework captures patterns of concern that are consistent with human reviewer reasoning while providing a baseline for evaluating human--AI alignment.

Unlike traditional classification tasks, where performance can be measured using metrics such as accuracy and F1-score, evaluating alignment between human and LLM-generated reviews requires assessing semantic correspondence between free-text review content. Although human evaluation is the gold standard and provides the most direct assessment of alignment, it is costly and difficult to scale to larger datasets. To investigate whether review alignment can be quantified, we employ Sentence-BERT (SBERT)~\cite{reimers2019sentence} and BERTScore~\cite{zhang2019bertscore} as semantic similarity measures. SBERT is used to assess category-level correspondence between human and LLM-generated review comments across the four evaluation categories, whereas BERTScore measures overall semantic similarity between corresponding review pairs and evaluates their ability to be distinguished from non-corresponding control pairs. The similarity scores produced by SBERT and BERTScore are compared with human ratings to assess how well these automated measures reflect the extent to which LLM-generated assessments capture concerns identified by human reviewers.
These automated analyses are intended to complement human evaluation and explore the feasibility of scalable assessment methods for future large-scale studies.

The contributions of this work are fourfold:
\begin{itemize}

\item We introduce intra-paper claim verification, a novel task for AI-assisted scientific peer review that evaluates whether novelty claims mentioned in the Introduction are substantiated by the methodological evidence presented within the same paper.
\item We develop a reviewer-inspired framework for operationalizing intra-paper claim verification. The framework derives evaluation categories from human reviewer feedback and uses them to structure review generation and evaluation, enabling systematic comparison between framework-generated assessments and human reviewer concerns.

\item We provide an empirical evaluation demonstrating that framework-generated assessments capture concerns that correspond to those raised by human reviewers.

\item We investigate SBERT-based semantic similarity measures as a scalable mechanism for automated evaluation of correspondence between framework-generated assessments and human reviewer feedback.
\end{itemize}

%% file: IEEE-RelatedWork.tex
\section{Related Work}
\label{sec:related-work}

\begin{figure*}[t] 
  \centering 
  \includegraphics[scale=0.5]{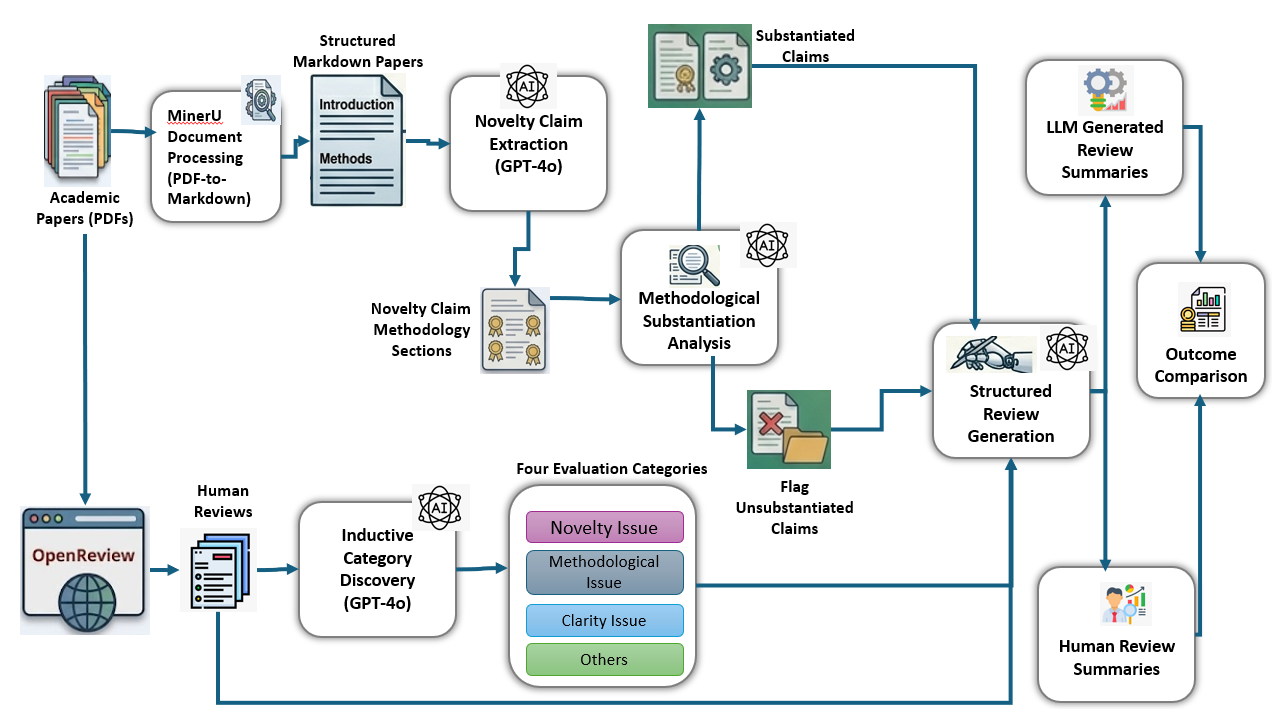} 
  \caption{Proposed framework for intra-paper claim verification and LLM peer review benchmarking. Academic papers are converted from PDF to structured markdown, from which novelty claims and methodology sections are extracted. A methodological substantiation analysis determines whether claimed contributions are supported by the methods section, identifying substantiated and unsubstantiated claims. These findings are incorporated into structured review generation. In parallel, human reviews are analyzed to derive four evaluation categories: Novelty Issue, Methodological Issues, Clarity Issues, and Other Issues. LLM-generated review summaries are subsequently compared with human review summaries to assess review alignment and benchmarking performance.} 
  \label{fig:Main} 
\end{figure*}

%\subsection{LLM-Assisted Peer Review}

Recent advancements in applying LLMs to the peer review pipeline focus primarily on engineering structured, automated critiques from historical review corpora. One prominent line of inquiry focuses on direct review generation. Frameworks such as REVIEWER2~\cite{gao2024reviewer2} and OpenReviewer~\cite{idahl2025openreviewer} fine-tune language models on extensive scientific review corpora to produce structured, aspect-conditioned critiques. To enhance feedback comprehensiveness and consistency, systems like AgentReview~\cite{jin2024agentreview} and MARG~\cite{d2024marg} simulate multi-agent reviewer deliberations to mirror real-world committee dynamics.

Rather than fully automating the generation process, an alternative paradigm focuses on targeted reviewer assistance. ReviewRobot~\cite{wang2020reviewrobot} leverages external retrieval mechanisms to find supporting evidence from related literature to ground evaluation claims. Similarly, MAMORX~\cite{taechoyotin2024mamorx} and AutoRev~\cite{chitale2025autorev} guide human reviewers by prompting coverage across essential evaluative criteria.

Recent research has increasingly examined novelty assessment as a distinct, specialized task within scientific peer review. Early studies investigated the role of document structure in identifying novelty-related signals, demonstrating that localized sections such as the \textit{Introduction}, \textit{Results}, and \textit{Discussion} contain stronger indicators of novelty than full-document representations~\cite{wu2025sc4anm}. These findings establish that section-aware analysis can improve automated novelty assessment by isolating the portions of a manuscript where authors explicitly articulate and contextualize their primary contributions.

Building on these structural insights, subsequent work has framed novelty assessment as an externally literature-grounded process. Afzal et al.~\cite{afzal-etal-2026-beyond} introduced a dedicated novelty-assessment pipeline that models human reviewer reasoning through systematic contribution extraction, related-work retrieval, and evidence-based comparison against a dynamically constructed research landscape. Similarly, Shahid et al.~\cite{shahid2025literature} proposed a retrieval-driven framework that identifies relevant prior work, filters and re-ranks candidate papers, and evaluates novelty through a comparative analysis of scientific contributions against the retrieved literature baseline.

Alternative methodologies perform fine-grained novelty analysis by extracting contribution claims and comparing them against structured representations of prior research. The OpenNovelty framework~\cite{zhang2026opennovelty} organizes retrieved literature into hierarchical taxonomies to conduct contribution-level comparisons supported by evidence extracted from scientific publications. Moving toward network-level representations, GraphMind~\cite{da-silva-etal-2025-graphmind} integrates citation networks, semantic retrieval, and structured paper decomposition to support evidence-based novelty exploration across the research landscape.

Unlike prior literature-grounded systems that evaluate a manuscript's novelty externally by comparing claims against prior work, the proposed framework focuses on a distinct axis of evaluation: intra-paper claim verification. This approach evaluates whether a manuscript provides sufficient internal evidence—specifically within its methodology—to substantively justify its own stated innovations. This internal verification addresses a critical blind spot in automated reviewer assistance; an entirely novel concept can pass external literature filters yet still fail to structurally support its core assertions. By targeting evidentiary sufficiency, this framework serves as a vital prerequisite that complements existing external retrieval tools.

%% file: IEEE-Methodology.tex
\section{Methodology}
\label{sec:method}

This study employs a four-stage pipeline, illustrated in Fig.~\ref{fig:Main}, to perform intra-paper claim verification and review generation. The framework first extracts novelty claims from the Introduction section of each manuscript and identifies methodological evidence relevant to those claims. Human peer reviews are then analyzed to inductively derive evaluation categories that reflect recurring reviewer concerns. Subsequently, the extracted claims and associated methodological evidence are used to perform claim--method substantiation analysis and generate structured LLM reviews using GPT-4o. In parallel, human reviewer comments are processed and organized according to the same evaluation categories, enabling direct comparison between human and LLM-generated assessments. The prompts used throughout the framework are publicly available in the accompanying GitHub repository.

\begin{figure}[t]
\vspace{-2mm}
    \centering
   \includegraphics[width=\columnwidth]{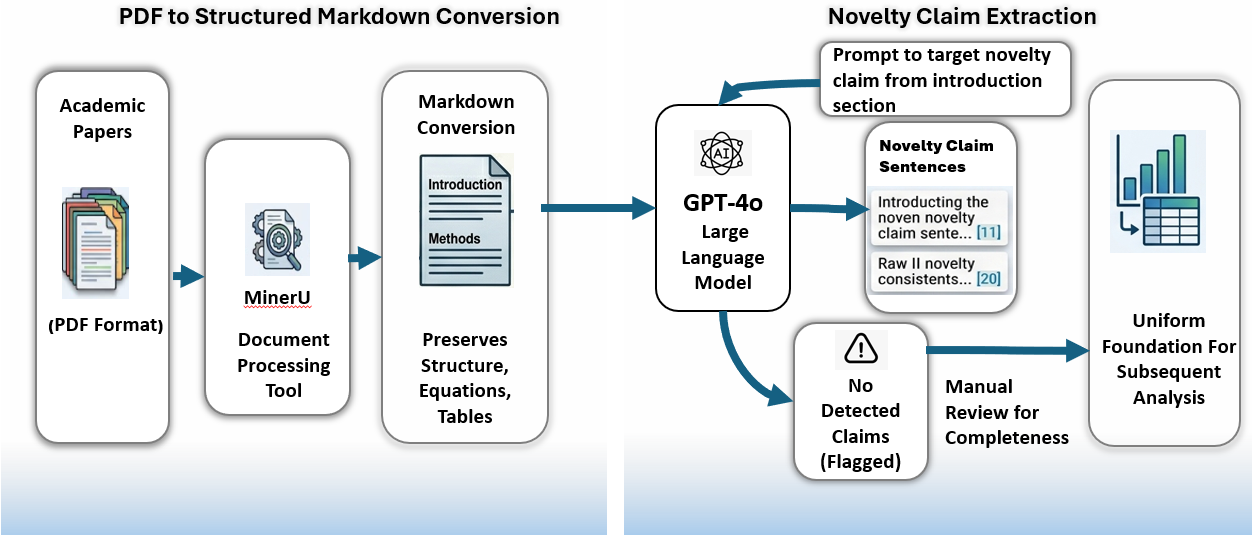}
    \caption{Workflow for PDF-to-Markdown conversion and novelty claim extraction using MinerU and GPT-4o}
    \vspace{-5mm}
    \label{fig:stage1}
\vspace{-2mm}
\end{figure}

\noindent \textit{Stage 1} identifies the novelty claims that constitute the objects of verification. In scientific papers, claims of innovation are typically introduced in the Introduction section, where authors describe the primary contributions and motivations of their work. Because the goal of this study is to determine whether these claimed innovations are substantiated by methodological evidence, the first step is to isolate the specific claims that will later be evaluated. Academic papers in PDFs are converted into machine-readable Markdown using MinerU~\cite{wang2024mineru}, after which novelty claims are extracted from the Introduction section using a fixed prompting strategy adopted from~\cite{afzal-etal-2026-beyond} as shown in Fig.~\ref{fig:stage1}. The output of this stage is a structured set of novelty claims that serves as the foundation for subsequent claim-verification analysis.

\begin{figure}[t]
\vspace{-2mm}
\centering
\includegraphics[width=\columnwidth]{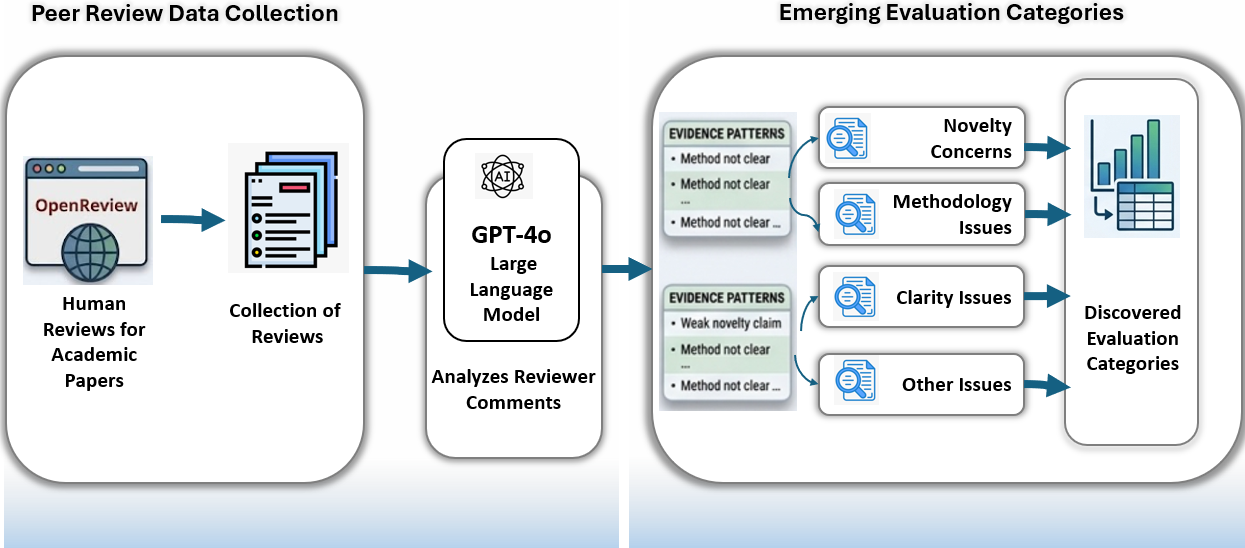}
%\vspace{-5mm}
\caption{Peer review analysis workflow for deriving evaluation categories from reviewer comments}
\vspace{-5mm}
\label{fig:stage2}
\vspace{-2mm}
\end{figure}

\noindent \textit{Stage 2} derives the evaluation categories used throughout the framework. Since reviewer concerns can vary substantially in focus and terminology, a common set of categories is required to organize assessments consistently across papers. Rather than relying on predefined dimensions, the categories are discovered inductively from human reviewer comments, allowing the framework to reflect evaluation criteria that naturally emerge during peer review, as shown in Fig.~\ref{fig:stage2}. Human reviews collected from OpenReview.net for 182 ICLR 2025 papers are analyzed to identify recurring patterns of reviewer concerns. The review corpus consists of 786 individual reviews (mean = 4.3 reviews per paper). Reviewer weaknesses and questions are first extracted from each review, then analyzed with GPT-4o to identify the distinct concerns raised by reviewers and generate concise concern labels. The resulting labels are then aggregated across the corpus and re-analyzed using GPT-4o to group semantically related concerns into higher-level themes. Examination of these recurring themes reveals four dominant categories of reviewer concerns: \textit{Novelty Issue}, \textit{Methodological Issue}, \textit{Clarity Issue}, and \textit{Other Issues}. Table~\ref{tab:category_example} presents representative reviewer comments corresponding to each of the derived evaluation categories. These categories provide a consistent structure for organizing the evidence-based assessments generated in subsequent stages of the pipeline.

\begin{itemize}
\item \textbf{Novelty Issue}: evaluates whether the proposed work demonstrates genuine novelty;
\item \textbf{Methodological Issue}: assesses the technical soundness and validity of the proposed approach;
\item \textbf{Clarity Issue}: examines the readability, organization, and comprehensibility of the paper;
\item \textbf{Other Issues}: captures additional concerns beyond the preceding categories, including issues related to scalability, generalization, and practical applicability.
\end{itemize}

\begin{figure}[!t]
\vspace{-2mm}
\centering
\includegraphics[width=\columnwidth]{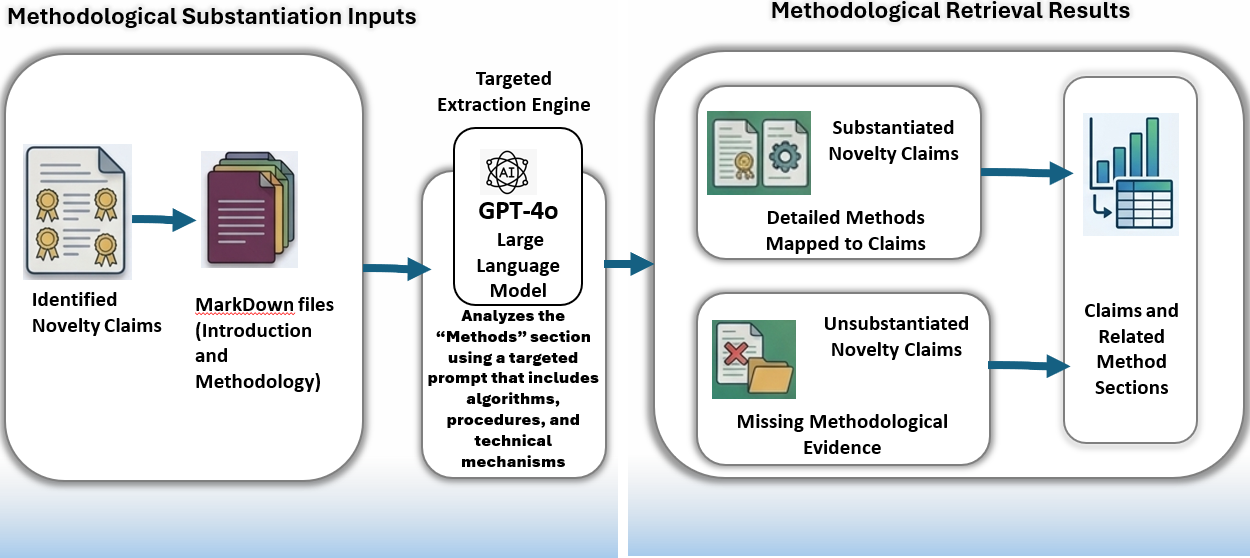}
%\vspace{-5mm}
\caption{Retrieval of methodological evidence for novelty claims}
\vspace{-5mm}
\label{fig:stage3}
\vspace{-2mm}
\end{figure}

\noindent\textit{Stage 3} performs intra-paper claim verification by retrieving methodological evidence corresponding to each novelty claim. Human reviewers frequently examine whether innovations described in the Introduction are concretely realized in the technical design presented later in the paper. A claim that cannot be traced to explicit methodological components may indicate insufficient substantiation, even when the claim itself appears novel. Accordingly, each extracted novelty claim is evaluated against the Methods section to identify supporting evidence such as algorithmic procedures, architectural modifications, training strategies, implementation details, or other technical mechanisms as shown in Fig.~\ref{fig:stage3}. The objective of this stage is to verify whether the paper internally provides methodological evidence supporting its stated innovations. Claims linked to explicit methodological support are considered substantiated, whereas claims lacking identifiable supporting evidence are flagged as potentially unsupported. 

\begin{table}[H]
\caption{Illustrative example of category discovery from reviewer comments.}
\label{tab:category_example}
\centering
\begin{tabular}{p{5.5cm} p{2.5cm}}
\hline
Reviewer Comment & Assigned Category \\
\hline
"The paper lacks novelty, as it appears to be an ensemble of SEED and EASE." & Novelty Concern \\

"A thorough comparison with these established methods is necessary to substantiate ALLoRA’s claims." & Methodological Issue \\

"The implementation details are insufficient for the reproduction." & Clarity Issue \\

"The proposed method may not scale to larger datasets." & Other Issues \\
\hline
\end{tabular}
\end{table}

\noindent\textit{Stage 4} translates the claim-verification findings into a structured review representation. The reviewer-derived categories established in Stage 2 provide a consistent framework for organizing assessments of claim substantiation. Using the novelty claims identified in Stage 1 and the methodological evidence retrieved in Stage 3, the framework generates category-specific review comments that articulate the strengths, limitations, and supporting evidence associated with each claimed contribution. This structured representation, as shown in Fig.~\ref{fig:stage4}, mirrors how reviewer feedback is commonly organized while maintaining explicit links between novelty claims and the methodological evidence supporting them.

To facilitate evaluation, human reviews are reorganized into the same four evaluation categories: Novelty, Methodology, Clarity, and Other issues. Because each paper may contain multiple reviewer reports, concerns expressed across all available reviews are extracted and consolidated within these categories. Both human and framework-generated reviews are subsequently summarized while preserving their original concerns, reasoning, and overall sentiment. Representing both review sources within a common structure reduces differences arising from writing style and organization, thereby enabling systematic comparison and downstream semantic similarity analyses.

\noindent\textit{Illustrative Example:} Consider the paper \textit{Self-Play with Execution Feedback: Improving Instruction-Following Capabilities of Large Language Models}. In the Introduction, the authors claim that their approach is the \textit{``first scalable and reliable method for automatically generating instruction-following training data.''} During Stage 1, the framework extracts this novelty claim. Stage 3 then retrieves supporting methodological evidence from the Methods section. For this claim, the extracted evidence describes AUTOIF, a framework that uses execution feedback from self-generated verification functions to supervise instruction-following behavior. The methods further explain how execution feedback, previously used in coding and tool-use tasks, is extended to natural language instructions through automatically generated verification functions and unit tests. This claim--method mapping is subsequently used as evidence for review generation.

Human reviews collected from OpenReview.net are initially available as unstructured reviewer comments containing sections such as strengths, weaknesses, and questions. For example, one reviewer noted that \textit{``Automatic instruction verification might not be an entirely new idea''} and referenced related efforts including CodeLlama2, PLUM, LLaMA3, SelfCodeAlign, and DeepSeek-Coder-V2. The framework categorizes this comment under the \textit{Novelty Concern} dimension and stores it as a structured reviewer observation.

Using the extracted novelty claim and corresponding methodological evidence, the framework generates a structured LLM review. For the selected claim, the generated assessment concludes that AUTOIF introduces a scalable approach based on execution feedback and self-generated verification functions. However, the review also notes that the broader concept of using execution feedback and automatically generated verification mechanisms is not entirely new, as related techniques have previously been explored in areas such as code synthesis and automated alignment. Consequently, the framework identifies a potential novelty concern regarding the degree of methodological differentiation from prior work.

Finally, both the categorized human reviews and the LLM-generated assessments are summarized. In this example, the human-reviewed summary highlights concerns about whether automatic instruction verification constitutes a genuinely novel contribution given prior work on execution-feedback-based alignment. Similarly, the LLM-review summary acknowledges the contribution of AUTOIF while noting that aspects of the underlying approach build upon existing execution-feedback paradigms. These summaries subsequently serve as inputs to evaluate the alignment between human and LLM-generated review assessments.

%% file: IEEE-Evaluation.tex
\section{Evaluation and Experiments}

\subsection{Validation of Intra-Paper Claim Verification}
\label{sec:validation}

This experiment evaluates the extent to which concerns generated by the proposed intra-paper claim verification framework align with those identified by human reviewers. The analysis is performed on a subset of the 182 ICLR 2025 papers~\cite{afzal-etal-2026-beyond} and their corresponding peer reviews collected from OpenReview.net. Human reviews were processed according to the methodology described in Section~\ref{sec:method}.

\begin{figure}[!t]
   \vspace{-2mm}
    \centering
    \includegraphics[width=\columnwidth]{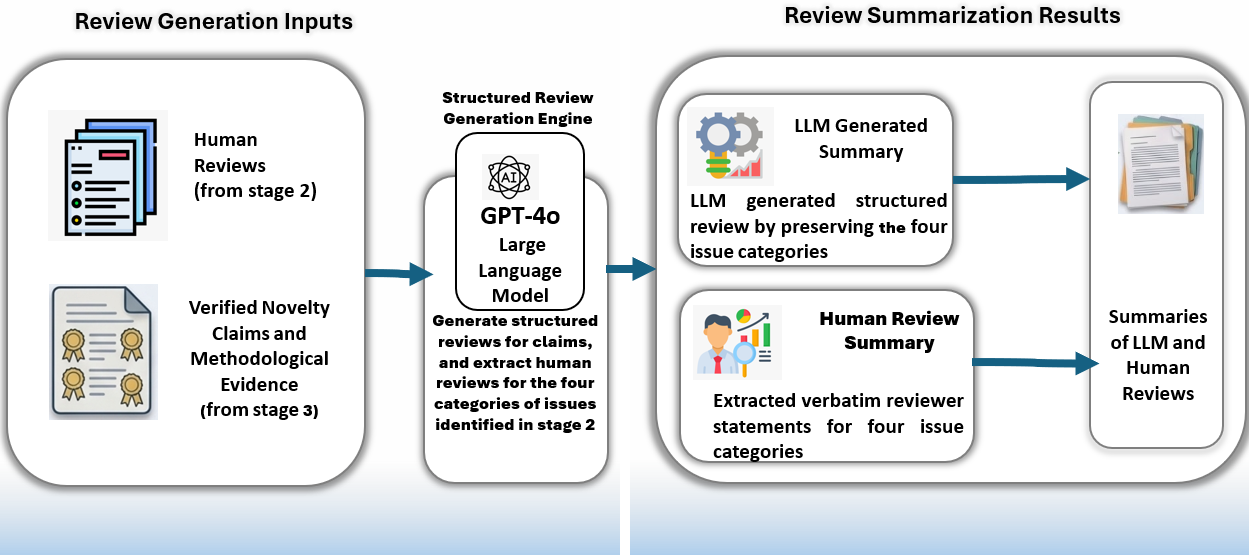}
    \caption{Structured review generation and summarization workflow}
    \vspace{-5mm}
    \label{fig:stage4}
    \vspace{-2mm}
\end{figure}

Twenty papers were randomly selected from the corpus, comprising 10 accepted and 10 rejected papers. This balanced selection captures a range of paper quality and reviewer concerns, enabling assessment of the framework across diverse review scenarios.

For each paper, both the human reviewed summary and the framework-generated summary were organized according to the four evaluation categories identified in Section~\ref{sec:method}. Two framework-generated reviews were then associated with each paper: a \textit{real} review generated from the target paper and a \textit{fake} review generated from a different paper in the larger dataset. Both reviews were produced using the same framework. To prevent information leakage, all fake reviews were drawn from papers outside the 20-paper evaluation subset while maintaining the same acceptance-outcome group (accepted or rejected). The fake reviews served as a control condition for evaluating whether reviewers could distinguish reviews generated from the target paper from those generated for unrelated papers. Presenting both review types in a blinded setting reduced potential evaluator bias and discouraged assumptions that every framework-generated review was necessarily relevant to the paper under evaluation.

The study was conducted by four scientists with expertise in artificial intelligence and machine learning and prior peer-review experience. To examine the influence of prior review experience, evaluators were partitioned into Evaluator Group 1 (Higher Review Experience) and Evaluator Group 2 (Lower Review Experience). Evaluators were first provided with the human review for a paper, which served as the reference description of reviewer concerns across the four evaluation categories. They subsequently evaluated \textit{Review A} and \textit{Review B} independently using a blinded protocol. Evaluators were not informed which review corresponded to the real framework-generated assessment and which represented the fake review.

Assessments were recorded using a five-point ordinal scale measuring the extent to which a review captured the same concerns identified by the human reviewer across the four evaluation categories. The detailed evaluation rubric is presented in Table~\ref{tab:1}.

This evaluation design enables direct assessment of whether the proposed framework identifies concerns similar to those raised by human reviewers while also verifying that the observed alignment is specific to the target paper rather than the result of review content.

\begin{table}[h]
\vspace{-3mm}
\caption{Rubric for evaluating LLM review assessment alignment with human reviews.}
\vspace{-3mm}
\begin{center}
\begin{tabular}{|c|p{7cm}|}
\hline
\textbf{Rating} & \textbf{Description} \\
\hline
1 & The LLM review raises completely different concerns from the human review — no meaningful overlap across any category \\
\hline
2 & The LLM assessment touches on some of the same concerns but misses the main issues the human raised \\
\hline
3 & The LLM review captures roughly half of the concerns the human reviewer identified across the four categories \\
\hline
4 & The LLM assessment captures most of the same concerns with only minor gaps \\
\hline
5 & The LLM assessment identifies the same concerns as the human review with no meaningful gaps \\
\hline
\end{tabular}
\label{tab:1}
\end{center}
\vspace{-7mm}
\end{table}

\subsection{Scalable Evaluation Using Automated Semantic Similarity Measures}
\label{sec:automated_eval}

Human evaluation provides the primary validation of the proposed framework because it directly measures whether framework-generated reviews identify concerns similar to those raised by human reviewers. However, human evaluation is time-consuming, costly, and difficult to scale to large collections of scientific papers. Consequently, we additionally investigate whether automated semantic similarity measures, namely SBERT and BERTScore, can serve as complementary proxies for expert assessment and provide a scalable mechanism for evaluating review alignment.

\textit{Sentence-BERT (SBERT)}~\cite{reimers2019sentence} is employed to measure semantic correspondence at the category level. Human and framework-generated review summaries are first organized into the four evaluation categories derived in Section~\ref{sec:method}: Novelty Issue, Methodological Issue, Clarity Issue, and Other Issue. SBERT generates dense sentence embeddings that capture semantic meaning beyond surface-level lexical similarity. Cosine similarity is then computed between the corresponding human and framework-generated category summaries. Because not every paper contained content for all evaluation categories, some category-level comparisons were unavailable. When either the human review summary or the framework-generated review summary for a category was absent, the corresponding comparison was excluded from the SBERT analysis rather than assigning a similarity score of zero. This category-level analysis enables examination of which review dimensions exhibit the strongest semantic agreement and provides a fine-grained view of alignment between framework-generated assessments and human reviewer concerns.

To determine whether the observed category-level similarity scores exceed a meaningful baseline, SBERT similarities are compared against a threshold of 0.50 using a one-sided Wilcoxon signed-rank test. The Wilcoxon test is selected because it does not assume normally distributed observations and is appropriate for the relatively small sample size of 20 papers. In addition to statistical significance, Cohen's $d$ is reported to quantify practical effect size.

\textit{BERTScore}~\cite{zhang2019bertscore} is employed as a document-level semantic similarity measure. Unlike SBERT, which evaluates correspondence within individual review categories, BERTScore assesses semantic similarity across the complete review summary. BERTScore leverages contextual token embeddings to identify semantic overlap even when reviewers express similar concerns using different wording or phrasing. Consequently, it provides an overall measure of review-level agreement between human and framework-generated assessments.

Together, these automated measures provide a scalable complement to human evaluation. Human ratings remain the primary indicator of alignment with expert reviewer reasoning, while SBERT and BERTScore offer additional evidence regarding category-level and document-level semantic correspondence between framework-generated assessments and human reviews.

%% file: IEEE-Result.tex
\vspace{-2mm}
\section{Results and Analysis}
\label{sec:results}
\vspace{-1mm}

\subsection{Human Evaluator Assessment}
\label{sec:performance-evaluation}

Four independent evaluators, comprising \emph{Evaluator Group 1} (Higher Review Experience) and \emph{Evaluator Group 2} (Lower Review Experience), assessed the correspondence between human reviews and LLM-generated reviews using a five-point ordinal rating scale. Across the 20 evaluated papers, the mean alignment score was 3.29, with a median of 3.50, as shown in Table~\ref{tab:reviewer_experience}. These results indicate moderate correspondence between the concerns identified by the proposed framework and those raised by human reviewers. While the framework captures a substantial portion of reviewer concerns, variability remains across individual papers and evaluation categories.

To examine the influence of prior review experience, partitioned evaluators, Evaluator Group 1 assigned a mean alignment score of 3.75, whereas Evaluator Group 2 assigned a mean score of 2.83. The difference between the two groups was statistically significant (paired $t=-5.40$, $p<0.001$; Wilcoxon $p<0.001$), indicating systematic differences in alignment ratings across evaluator groups.

Alignment scores were further analyzed separately for accepted and rejected papers. Evaluator Group 1 assigned comparable scores to accepted papers -- 3.60 and rejected papers -- 3.90. In contrast, Evaluator Group 2 assigned lower scores overall, with the lowest mean score observed for rejected papers -- 2.60. These results indicate that perceptions of alignment varied across evaluator groups, particularly for rejected submissions.

\begin{figure}[t]
%\vspace{-3mm}
    \centering
    \includegraphics[width=0.48\textwidth]{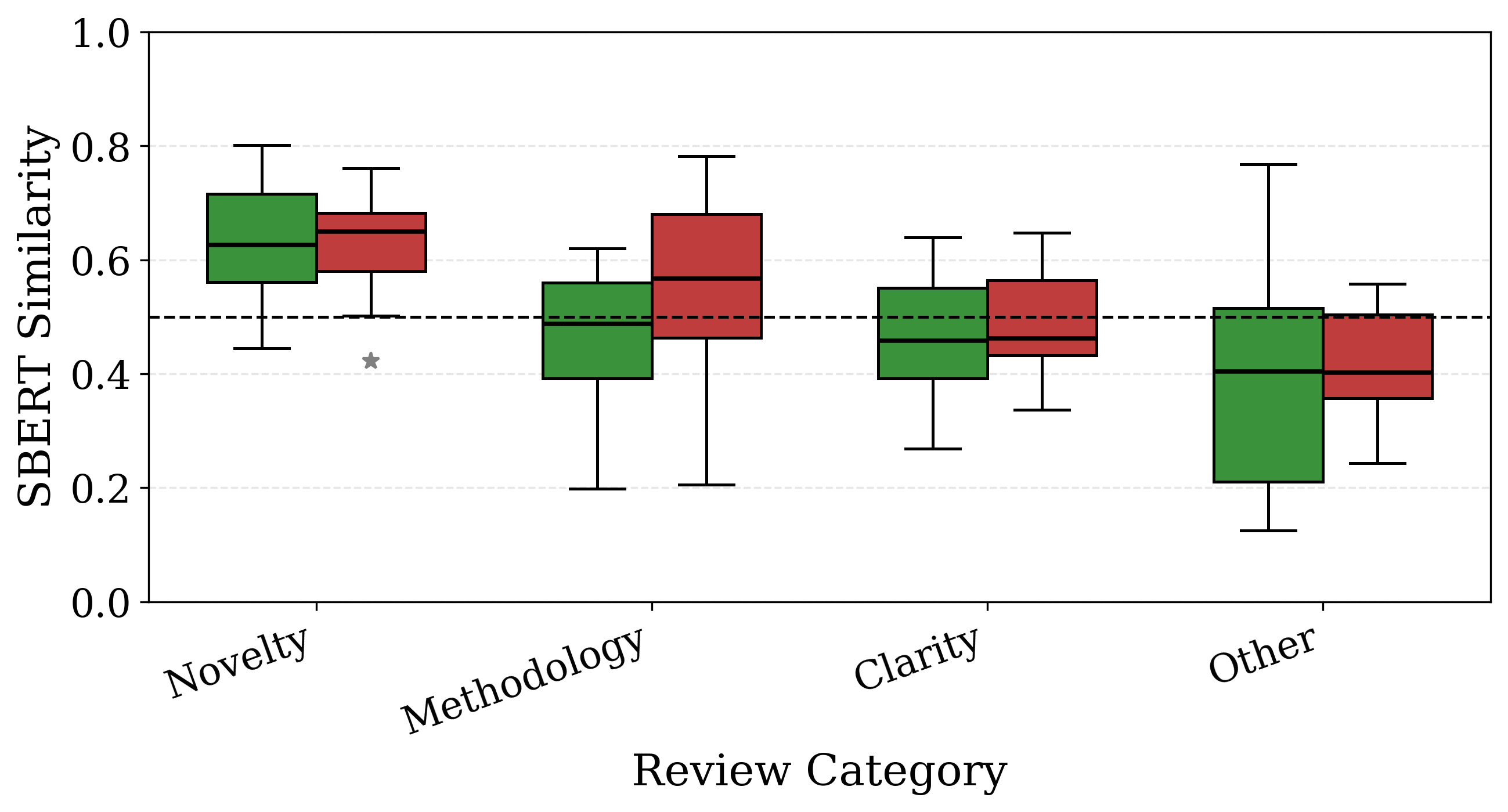}
   % \vspace{-2mm}
    \caption{Category-level SBERT similarity distributions for accepted (green) and rejected (red) papers. The dashed line indicates the similarity baseline of 0.50.}
    \label{fig:acceptedRejected}
   % \vspace{-5mm}
\end{figure} 
The higher alignment scores assigned by Evaluator Group 1 for rejected papers suggest stronger recognition of correspondence between concerns identified by human reviewers and those generated by our proposed framework. Rejected papers typically contain more explicit reviewer criticisms, and the results indicate that evaluators with greater review experience perceived a higher degree of agreement between the two sources of assessment.

\begin{table}[H]
\caption{Alignment scores by reviewer experience and paper outcome.}
\label{tab:reviewer_experience}
\centering
\begin{tabular}{lcc}
\hline
\textbf{Metric} & \textbf{Evaluator Group 1} & \textbf{Evaluator Group 2} \\
Mean Alignment Score & 3.75 & 2.83 \\
Standard Deviation & 0.38 & 0.63 \\
Accepted Papers & 3.60 & 3.05 \\
Rejected Papers & 3.90 & 2.60 \\
\hline
\end{tabular}
%\vspace{-5mm}
\end{table}

\subsection{Category-Level Semantic Alignment}

\begin{figure}[H]
    \centering
    \includegraphics[width=0.4\textwidth]{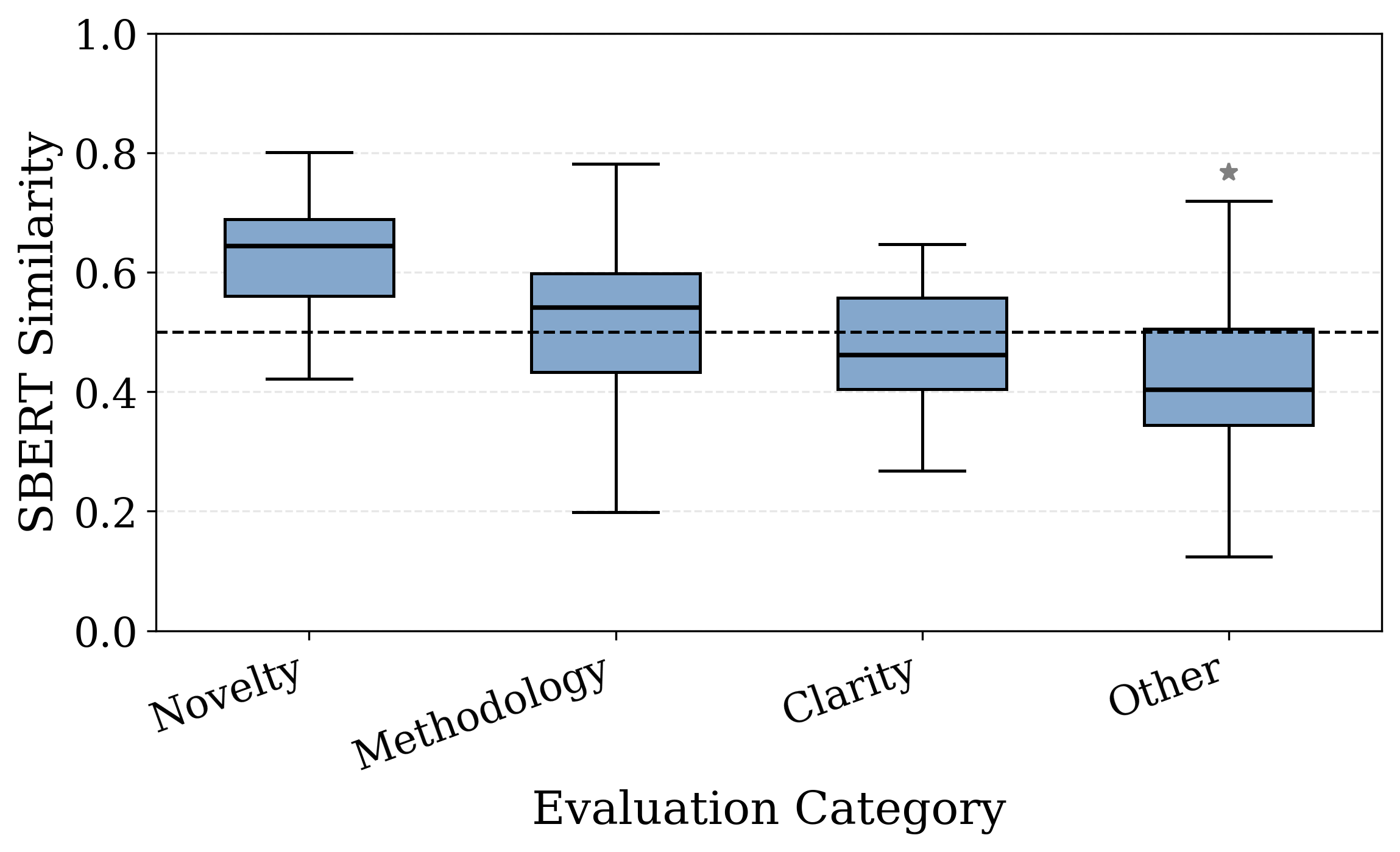}
    \vspace{-2mm}
    \caption{Category-level SBERT similarity distributions across novelty, methodology, clarity, and other review dimensions. The dashed line indicates the similarity baseline of 0.50.}
    \label{fig:CategorySBERT}
    \vspace{-2mm}
\end{figure}

To examine alignment across specific review dimensions, SBERT cosine similarity was computed independently for the four evaluation categories: Novelty, Methodology, Clarity, and Other issues. The resulting distributions are shown in Fig.~\ref{fig:CategorySBERT}, while summary statistics are reported in Tables~\ref{tab:sbert_similarity} and \ref{tab:sbert_significance}.The number of observations varied across categories. Novelty, Methodology, and Clarity each contained 20 paper-level comparisons, whereas the Other category contained 18 comparisons because two papers lacked corresponding content in this category and were therefore excluded from the analysis.

As shown in Table~\ref{tab:sbert_similarity}, Novelty achieved the highest mean similarity score (0.628), followed by Methodology (0.508), Clarity (0.472), and Other (0.415). Novelty was the only category consistently exceeding the baseline similarity threshold of 0.50 and exhibited the strongest semantic correspondence between human and LLM-generated reviews. This may reflect the fact that novelty-related concerns are often stated more explicitly and directly in both human reviews and framework-generated assessments. In contrast, methodological and clarity-related concerns could be expressed in more diverse ways, leading to lower similarity scores. However, further investigation is required to better understand the factors contributing to these differences across evaluation categories.

Statistical analysis confirmed that Novelty was the only category demonstrating significant improvement above the baseline threshold (Wilcoxon $p < 0.001$, Cohen's $d = 1.17$). Although Methodology achieved a mean similarity slightly above the baseline, the difference was not statistically significant. Clarity and Other exhibited lower similarity scores and failed to exceed the baseline threshold.

When analyzed separately by paper outcome, accepted and rejected papers exhibited largely overlapping similarity distributions across all categories, as shown in Fig.~\ref{fig:acceptedRejected}. Novelty remained the highest-alignment category for both groups, while Methodology and Clarity demonstrated moderate similarity and Other exhibited the lowest similarity scores. The substantial overlap between accepted and rejected distributions suggests that semantic alignment is influenced more by the type of reviewer concern than by the final acceptance decision.

Overall, these findings indicate that novelty-related concerns show the strongest correspondence between human and LLM-generated reviews and that this pattern remains consistent across both accepted and rejected papers.

\begin{table}[H]
\vspace{-2mm}
\caption{Category-level SBERT similarity statistics.}
\label{tab:sbert_similarity}
\centering
\begin{tabular}{lccc}
\hline
\textbf{Category} & \textbf{N} & \textbf{Mean Similarity} & \textbf{SD} \\
\hline
Novelty     & 20 & 0.628 & 0.110 \\
Methodology & 20 & 0.508 & 0.158 \\
Clarity     & 20 & 0.472 & 0.107 \\
Other       & 18 & 0.415 & 0.170 \\
\hline
\end{tabular}
\vspace{-2mm}
\end{table}

\subsection{Discriminative Capability of BERTScore}

While SBERT evaluates category-level semantic alignment, it does not assess whether a generated review corresponds to the correct paper. Therefore, BERTScore was used to compare human reviews against both corresponding LLM-generated reviews and non-corresponding reviews generated for different papers.

\begin{figure}[H]
    \centering
    \includegraphics[width=0.3\textwidth]{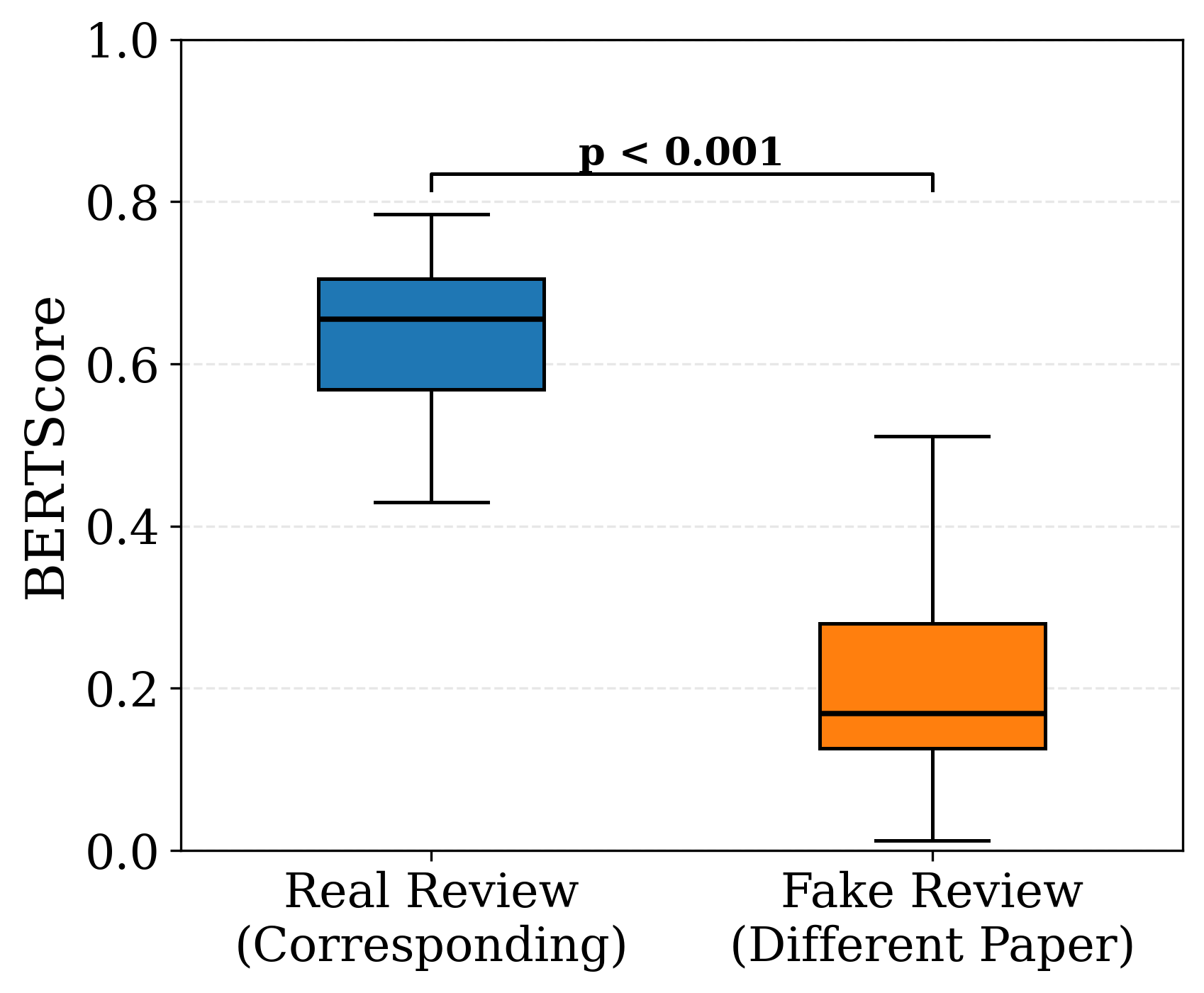}
    \caption{Distribution of BERTScore similarity between human reviews and corresponding LLM-generated reviews (Real) compared with non-corresponding reviews from different papers (Fake).}
    \label{fig:BertScoreRealFake}
  %  \vspace{-5mm}
\end{figure}

As shown in Fig.~\ref{fig:BertScoreRealFake}, corresponding reviews consistently achieved substantially higher BERTScore values than non-corresponding reviews. Real reviews outperformed fake reviews for all evaluated papers, yielding a discrimination rate of 100\%.

A paired $t$-test confirmed that the difference between the two distributions was statistically significant ($p < 0.001$). These results suggest that BERTScore may capture review-specific semantic correspondence and be sensitive to the relationship between a review and its target paper.

\begin{table}[t]
\caption{Statistical comparison of category-level SBERT similarity against the baseline threshold of 0.50.}
\label{tab:sbert_significance}
\centering
\begin{tabular}{lcccc}
\hline
\textbf{Category} &
\textbf{One-Sided} &
\textbf{Wilcoxon} &
\textbf{Wilcoxon} &
\textbf{Cohen's} \\
&
\textbf{$t$-test $p$} &
\textbf{Statistic} &
\textbf{$p$-value} &
\textbf{$d$} \\
\hline
Novelty     & $<0.001$ & 198.0 & $<0.001$ & 1.17 \\
Methodology & 0.413    & 116.0 & 0.351    & 0.05 \\
Clarity     & 0.871    & 74.0  & 0.877    & -0.26 \\
Other       & 0.976    & 39.0  & 0.981    & -0.50 \\
\hline
\end{tabular}
\vspace{-5mm}
\end{table}

%\subsection{Tools and Implementation}

%MinerU \cite{wang2024mineru} was used for PDF-to-Markdown conversion, GPT-4o accessed through the OpenAI API was used for novelty claim extraction, category discovery, claim--method verification, and structured review generation, and OpenReview \cite{openreview} was used for collecting human reviewer comments. Category-level semantic similarity was computed using SBERT embeddings generated by the \texttt{all-mpnet-base-v2} model implemented through the \texttt{sentence-transformers} library. Overall semantic similarity between human and LLM-generated reviews was measured using the \texttt{bert-score} package.

%All data processing and statistical analyses were conducted in Python. The \texttt{pandas} and \texttt{NumPy} libraries were used for data manipulation and numerical computation. Statistical analyses, including one-sample hypothesis testing, paired $t$-tests, Wilcoxon signed-rank tests, and correlation analyses, were performed using \texttt{SciPy}. Visualization and figure generation were implemented using \texttt{Matplotlib} and \texttt{Seaborn}. These tools collectively supported data preparation, semantic similarity computation, statistical evaluation, and result visualization throughout the study.

%% file: IEEE-LimitationConclusion.tex
\section{Discussion}

%The proposed framework evaluates whether novelty claims are supported by methodological evidence within a paper, providing complementary information to literature-based novelty assessment for peer-review decision-making.
The results demonstrate that the proposed framework can identify concerns that correspond to those raised by human reviewers by linking novelty claims to supporting methodological evidence, providing a structured mechanism for assessing whether scientific claims are adequately substantiated.

The evaluator experience analysis further provides insight into how framework-generated assessments are perceived. Evaluators with greater reviewing experience consistently identified stronger correspondence between framework-generated and human reviewer concerns. One possible explanation is that experienced evaluators are more accustomed to reasoning about the relationship between scientific claims, supporting evidence, and reviewer criticism. This finding suggests that the framework may capture aspects of reviewer reasoning that are more readily recognized by experienced evaluators. If this interpretation holds, AI-assisted review systems may serve not only as tools for automating aspects of peer review, but also as mechanisms for exposing less experienced reviewers to the evaluative strategies commonly employed by experienced reviewers. Such systems could support reviewer training by making more explicit the connections between scientific claims, methodological evidence, and the concerns that arise when those connections are weak.

This observation has implications beyond the specific task examined in this study. Rather than viewing LLMs solely as tools for automating peer review, future research may investigate how reviewer-derived reasoning frameworks can support reviewer decision-making, reviewer training, and greater consistency across evaluations. Similar approaches may also be extended to other dimensions of scientific assessment, including methodological rigor, reproducibility, significance, and experimental design. More broadly, the findings suggest that combining LLMs with structured representations of expert evaluation practices may provide a promising direction for developing AI systems that support complex scientific reasoning tasks.

\section{Limitations and Future Work}
\label{sec:limitations}
The evaluation was performed on a subset of 20 papers drawn from a corpus of 182 ICLR 2025 submissions, limiting the generalizability of the findings. In addition, the evaluation categories were derived from reviewer feedback within this corpus and may not fully reflect evaluation practices in other research communities or publication venues. While novelty exhibited the strongest semantic correspondence between human and LLM-generated assessments, methodology and clarity demonstrated only moderate similarity, suggesting that these dimensions may be more challenging to capture consistently through automated review generation. Future work will extend the framework to additional conferences and scientific disciplines and explore integrating literature-grounded novelty assessment with claim-substantiation analysis to provide a more comprehensive evaluation of scientific contributions.

\section{Conclusion}
\label{sec:conclusion}

This paper introduced intra-paper claim verification, a new task for AI-assisted scientific peer review that evaluates whether a manuscript’s methodological evidence substantiates its claimed contributions. It complements existing literature-grounded novelty assessment, which evaluates whether claimed contributions differ from prior work.

Using reviewer-derived evaluation criteria and structured claim–method analysis, the framework generated assessments that demonstrated meaningful alignment with concerns raised by human reviewers, particularly for novelty-related issues. We additionally explored automated semantic similarity measures as a potential mechanism for scaling future evaluation beyond the limits of expert human assessment.

These findings suggest that structured representations of reviewer reasoning can be used to guide LLM-based scientific evaluation, creating opportunities for AI-assisted review, reviewer support, and more consistent assessment of scientific contributions.

\section*{Acknowledgment}
This research used the Bridges-2 system at the Pittsburgh Supercomputing Center (PSC) to support large-scale processing and analysis of scientific papers. The manuscript was improved and proofread with assistance from Grammarly. Large language models were used  during the development and evaluation of the intra-paper claim verification framework. All LLM-generated content was carefully reviewed, verified, and edited by the authors.